\documentclass[11pt]{article}

\usepackage[final]{acl}

\usepackage{times}
\usepackage{latexsym}

\usepackage[T1]{fontenc}

\usepackage[utf8]{inputenc}

\usepackage{microtype}

\usepackage{inconsolata}

\usepackage{graphicx}

\usepackage{tcolorbox}
\tcbuselibrary{skins}
\usepackage[dvipsnames]{xcolor}
\usepackage{enumitem}
\usepackage{booktabs}
\usepackage{xspace}

\definecolor{lightgray}{HTML}{F5F5F5}
\definecolor{bordergray}{HTML}{BBBBBB}
\definecolor{labelgray}{HTML}{666666}

\newcommand{\dsetname}{\textsc{WikiVideo25}\xspace}

\title{Findings of the MAGMaR 2026 Shared Task}

\author{
\textbf{Alexander Martin}\textsuperscript{\rm 1}
\quad \textbf{Dengjia Zhang}\textsuperscript{\rm 1}
\quad \textbf{Joel Brogan}\textsuperscript{\rm 2}\thanks{Subsequent authors listed alphabetically}
\quad \textbf{Francis Ferraro}\textsuperscript{\rm 3}
\\
\textbf{Jeremy Gwinnup}\textsuperscript{\rm 4}
\quad \textbf{Reno Kriz}\textsuperscript{\rm 5}
\quad \textbf{Teng Long}\textsuperscript{\rm 6}
\\
\textbf{Kenton Murray}\textsuperscript{\rm 5}
\quad \textbf{Andrew Yates}\textsuperscript{\rm 5}
\quad \textbf{Xiang Xiang}\textsuperscript{\rm 7}
\\
\textsuperscript{1}Johns Hopkins University
\quad \textsuperscript{2}OpenAI
\\
\textsuperscript{3}University of Maryland, Baltimore County
\quad \textsuperscript{4}Air Force Research Laboratory
\\
\quad \textsuperscript{5}Human Language Technology Center of Excellence, Johns Hopkins University
\\
\textsuperscript{6}University of Amsterdam
\quad \textsuperscript{7}Huazhong University of Science and Technology
\\
\texttt{\small{\{amart233, kenton, rkriz1\}@jhu.edu}}
}

\begin{document}
 \maketitle
\begin{abstract}
This overview paper presents the results of the shared task for the second workshop on Multimodal Augmented Generation via Multimodal Retrieval (MAGMaR). In this shared task participants submitted systems focused on either (i) video retrieval or (ii) grounded generation of articles given retrieved videos. Teams could submit to either task. For the retrieval task, we had 2 participating teams that submitted a total of 17 systems -- all of which beat a baseline derived from the winner of last year's shared task. On the generation side, we had 4 teams submit 16 systems. All teams had at least one generated report that was labeled the best by a human annotator.

\end{abstract}

\section{Introduction}
The second workshop on Multimodal Augmented Generation via Multimodal Retrieval (MAGMaR 2026) was located at ACL in San Diego and hosted a shared task focused on video retrieval and generation. Increasingly, information seeking needs require solutions that span modalities, yet much of the literature, and open shared tasks, continue to focus on one modality (i.e., text).  The focus of this shared task is aimed at addressing this deficiency.
The task built upon the success of last year's shared task of retrieving videos given a query. This year the task expanded the test collection of videos and introduced a new generation track:
\begin{itemize}
    \item \textbf{Retrieval Track:} Systems provide a ranked list of videos in the collection ordered by relevance to the query.
    \item \textbf{Generation Track:} Systems produce a text article that answers the information need and grounds its content in the retrieved videos.
\end{itemize}

This shared task focuses on retrieving relevant videos and generating grounded articles that respond to information needs. Given a query describing a real-world current event, participating systems must identify pertinent videos from a large multilingual, multimodal collection and use that evidence to produce a coherent and informative written article. Our retrieval collection comprised over 110,000 multilingual, event-centric videos. The generation task had 19 queries for which systems needed to produce an article that met the information seeking needs of a specific persona. The summaries were judged by 3 human participants.

Overall, this was a very successful shared task garnering multiple submissions with all teams beating very strong baselines in both tracks. Teams made meaningful improvements and explored cutting-edge solutions to an open-problem. We are delighted with the results and suggest you read the system descriptions. That being said, here are some of the highlights across submissions:

\paragraph{Findings of the 2026 MAGMaR Shared Task}
\begin{itemize}
    \item Text-based reasoning dominated both tracks. Converting videos into summaries and captions rather than operating on video directly helps in both retrieval and generation.
    \item Complex, persona-constrained queries issued against a large video corpus do not perform well for retrieval methods using only a single dense embedding. However, more expressive retrieval architectures can do substantially better.
    \item The best performing retrieval systems relied on broad first-stage recall with reasoning-based reranking and significantly outperformed all baselines.
    \item There is a clear disconnect between human preference and automatic metrics for scoring generations. Systems that abstain when evidence is insufficient are rewarded by annotators yet penalized by the automatic metric as precision failures.
\end{itemize}

\section{Data}
\begin{figure*}
    \centering
    \includegraphics[width=\linewidth]{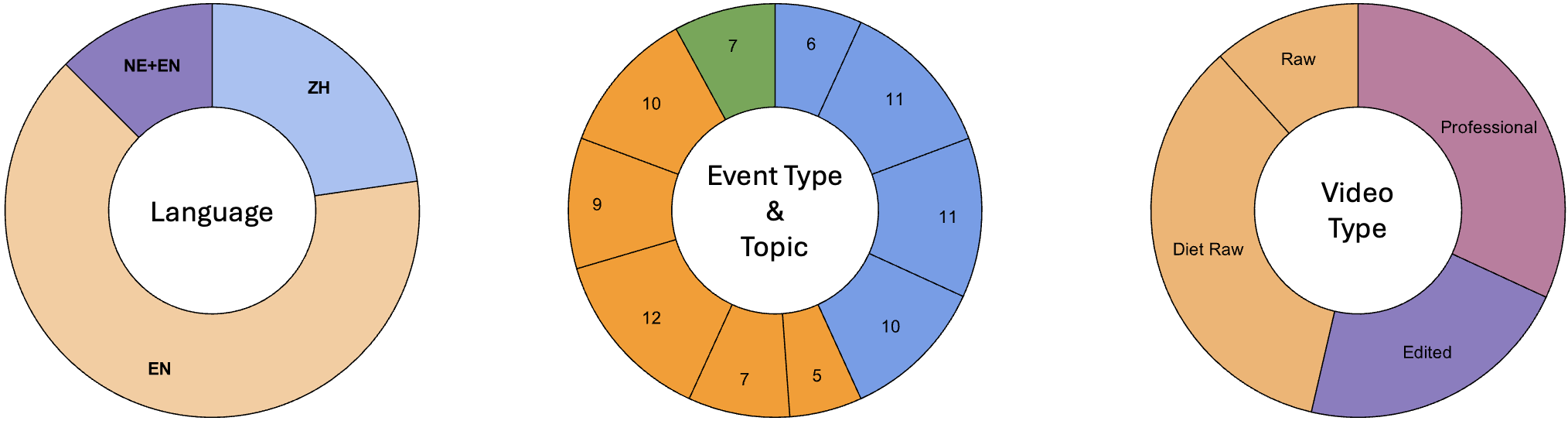}
    \caption{\dsetname Summary Statistics. ``Language'' (left): Videos are in English, Chinese, and Nepali. ``Event Type \& Topic'' (center): are partitioned by number of videos per topic and color coded by event type. Event Type Key: \textcolor{orange}{Orange}: Natural Disasters, \textcolor{NavyBlue}{Blue}: Political Developments, \textcolor{ForestGreen}{Green}: Science. ``Video Type'' (right): Nearly half of videos are raw and diet raw footage, captured by bystanders without professional editing or narration.}
    \label{fig:stats}
\end{figure*}

\begin{figure*}[t]
\centering
\small
\begin{tcolorbox}[
  colback=lightgray, colframe=bordergray,
  boxrule=0.5pt, arc=1.5pt,
  left=7pt, right=7pt, top=6pt, bottom=6pt,
  width=\textwidth
]
 
\textbf{Query 1}\\[3pt]
\textcolor{labelgray}{\textsc{Persona}}\quad
Statistician for North American Elections\\[3pt]
\textcolor{labelgray}{\textsc{Query}}\quad
Help me compile parliamentary and vote share statistics on the 2025
Canadian Federal elections. This should include how many seats each
party won or lost, how many total seats each party now has, any
information on total (popular) vote share available for the major
parties and for any specific candidates for which it is available.
Any demographic breakdowns of support for particular parties or
candidates would also be helpful. For all statistics, please also
include as much detail as possible on their source, as this helps me
get a sense for how credible they are.
 
\vspace{5pt}
\noindent\textcolor{bordergray}{\rule{\linewidth}{0.4pt}}
\vspace{5pt}
 
\textbf{Query 2}\\[3pt]
\textcolor{labelgray}{\textsc{Persona}}\quad
Policy Analyst at Elections Canada\\[3pt]
\textcolor{labelgray}{\textsc{Query}}\quad
I am looking to understand the key dynamics and outcomes of the 2025
Canadian Federal Election. Specifically, I am interested in how vote
shares translated into seat outcomes under the first-past-the-post
system, the regional patterns that enabled the Liberals to form
government, and the factors contributing to the sharp decline in NDP
support. Information on polling trends during the campaign, the role
of leadership changes, and the influence of external political
pressures---such as U.S.\ political rhetoric---would be particularly
valuable.
 
\end{tcolorbox}
 
\caption{Two example queries from \textsc{WikiVideo25} for the
  \textit{2025 Canadian Federal Election} event. Query~1 is
  \textit{unbiased} and adopts a statistician persona seeking
  numerical data; Query~2 is \textit{biased} and adopts a policy
  analyst persona focused on structural explanations. Both target
  English-language output.}
\label{fig:query-example}
\end{figure*}

The shared task is built on \textsc{WikiVideo25}, which extends \textsc{WikiVideo} \cite{martin2025wikivideoarticlegenerationmultiple} to address its limitations in query specificity, language coverage, video diversity, and annotation methodology. \autoref{fig:stats} gives a general visualization, and \autoref{append:data} provides a detailed comparison with the original dataset.

\paragraph{Queries and personas.}
Each topic comes with a persona and a query (two per topic; see \autoref{fig:query-example}). The persona constrains what counts as relevant evidence, so systems must reason about which parts of the videos matter for a specific professional need. Each query is written in one of two ways: \textit{biased}, where the author had watched the videos, and \textit{unbiased}, where the author had only seen video titles. Unbiased queries may ask for information absent from the collection, testing whether systems can acknowledge gaps rather than hallucinate.

\paragraph{Annotation and video diversity.}
Ground-truth articles are built bottom-up: annotators first write atomic claims grounded in each video's audiovisual content, then combine them into higher-level inferences. This produces reports faithful to the videos and reduces the advantage of parametric knowledge. The video collection shifts toward raw and ``diet raw'' footage, videos captured by bystanders without professional narration, which force models to infer events from low-level audiovisual cues instead of relying on transcripts alone. \textsc{WikiVideo25} also adds two Mandarin-language topics and one mixed Nepali--English topic, so systems must process non-English audio and OCR.

\paragraph{MultiVENT 2.0.} 
Last year's shared task only had the retrieval track. We used the test collection of MultiVENT 2.0 \citep{kriz2025multivent20massivemultilingual}, a dataset of over 110,000 videos across 6 languages. This year, these videos were used as a distractor set which was combined with \textsc{WikiVIDEO25} to form the test collection. The videos in MultiVENT2.0 and \textsc{WikiVideo25} come from similar sources, video types, and distributions. Because MultiVENT 2.0 covers events only through 2023 and all \textsc{WikiVideo25} events occur in 2025, no distractor video can be a true positive for any query; relevance judgments therefore apply exclusively to the \textsc{WikiVideo25} portion of the collection.

\section{Retrieval Task}
\begin{table*}[ht]
\centering
\small
\begin{tabular}{@{}clcccccc@{}}
\toprule
\textbf{Rank} & \textbf{System}
	& \textbf{nDCG@10} & \textbf{nDCG@20} & \textbf{nDCG@100}
	& \textbf{R@10} & \textbf{R@20} & \textbf{R@100} \\
\midrule
-- & \textit{OmniEmbed}
	& \textit{0.166} & \textit{0.186} & \textit{0.245} & \textit{0.096} & \textit{0.158} & \textit{0.297} \\
-- & \textit{OmniEmbed + RankVideo}
	& \textit{0.542} & \textit{0.534} & \textit{0.546} & \textit{0.423} & \textit{0.462} & \textit{0.494} \\
-- & \textit{Mixedbread}
	& \textit{0.717} & \textit{0.706} & \textit{0.748} & \textit{0.604} & \textit{0.634} & \textit{0.741} \\
\midrule
6 & MARQUIS-6
	& 0.746 & 0.757 & 0.799 & 0.636 & 0.711 & 0.818 \\
5 & MARQUIS-8
	& 0.747 & 0.758 & 0.800 & 0.636 & 0.711 & 0.818 \\
4 & MARQUIS-4
	& 0.754 & 0.765 & 0.807 & 0.641 & 0.716 & 0.823 \\
3 & MARQUIS-14
	& 0.757 & 0.768 & 0.810 & 0.650 & 0.725 & 0.832 \\
2 & MARQUIS-2
	& 0.759 & 0.771 & 0.811 & 0.652 & 0.730 & 0.832 \\
\textbf{1} & \textbf{C2F-RAG}
	& \textbf{0.848} & \textbf{0.851} & \textbf{0.853} & \textbf{0.773} & \textbf{0.832} & \textbf{0.837} \\
\bottomrule
\end{tabular}
\caption{MAGMaR 2026 --- Retrieval track leaderboard (sorted by nDCG@10). Baselines shown in italics above the rule. Note that we received 17 submission and all of them beat a strong baseline based on the winner of last year's shared task. As many submitted systems were slight variants of each other, we are only showing the top 6 systems from the leaderboard.}
\label{tab:retrieval}
\end{table*}

\paragraph{Task.}
The retrieval task requires systems to rank a corpus of over 110,000 multilingual, event-centric videos by relevance to a structured query. As illustrated in \autoref{fig:query-example}, each query comes with a detailed persona and a complex information need. Systems must return a ranked list of the top 1000 videos and are evaluated by nDCG and Recall at cutoffs of 10, 20, and 100, measured using \texttt{ir-measures} \cite{macavaney2022}, over the combined \textsc{WikiVideo25} and MultiVENT 2.0 \cite{kriz2025multivent20massivemultilingual} test corpora.

\subsection{Systems}
We provide participants with three baselines: OmniEmbed \cite{ma2025tevatron20unifieddocument}, an open-source dense retriever; OmniEmbed + RankVideo \cite{skow2026rankvideoreasoningrerankingtexttovideo}, OmniEmbed reranked by an open-source video-native reranker; and Wholembed-v3 \cite{mixedbread2026wholembed}, a closed-source first-stage retrieval model.

We received 17 submissions\footnote{Only showing the top 5 from MARQUIS} from two systems: C2F-RAG~\cite{dai26} and MARQUIS~\cite{chakraborty2026marquisthreestagepipelinevideo}. Both follow a similar pipeline: broad first-stage recall followed by reasoning-based reranking, with no task-specific training. C2F-RAG indexes text proxies of the video content (global summaries and keyframe captions) with BGE-M3 \cite{chen2025m3embeddingmultilingualitymultifunctionalitymultigranularity}, then applies an LLM-based cognitive reranking agent that scores each candidate on logical alignment with the query persona. MARQUIS operates on the video content directly, decomposing each query into atomic sub-queries, retrieving independently over each with OmniEmbed, fusing the resulting ranked lists, and reranking with RankVideo.

\subsection{Results}

\autoref{tab:retrieval} summarizes the retrieval results.

\paragraph{First-stage recall vs.\ reasoning-based reranking.}
The baselines span a wide performance range. OmniEmbed achieves an
nDCG@10 of 0.17, which shows how difficult it is to match complex,
persona-constrained queries against a large video corpus with a single
embedding. Adding RankVideo as a reranker more than triples this to
0.54; even over a weak first stage, a reasoning-based second stage
makes a large difference. Wholembed-v3, a closed-source late-interaction
model, reaches 0.72 without any reranking, so more expressive retrieval
architectures can close much of the gap on their own. Both submitted
systems pair broad first-stage recall with reasoning-based reranking
and outperform all baselines.

\paragraph{Text proxies vs.\ video-native retrieval.}
C2F-RAG achieves the highest scores on every metric (nDCG@10 of 0.848)
despite never operating on video content directly; it indexes text
proxies (global summaries and keyframe captions) and reranks with an
LLM-based cognitive agent. MARQUIS operates on video directly through
OmniEmbed, yet its best configuration (MARQUIS-2) reaches only 0.759.
Whether text proxies are a better retrieval substrate for complex
queries than current video embeddings, or whether any reasonable first
stage would perform similarly when paired with a strong reranker,
remains an open question.

\paragraph{Query decomposition and rank fusion.}
MARQUIS decomposes each query into atomic sub-queries, retrieves
independently over each, and fuses the resulting ranked lists before
reranking. Starting from OmniEmbed, the weakest first-stage baseline
(0.17), this strategy closes the gap: MARQUIS-2 reaches 0.759. The
tight clustering of MARQUIS variants (0.746--0.759), however, suggests
that the gains come mainly from query expansion and reranking, not from
the choice of fusion strategy.

Several questions remain. Combining a stronger first-stage model like
Wholembed-v3 with C2F-RAG's cognitive reranking, or with a trained
video-native reranker, would help separate the contributions of recall
from reasoning. More broadly, the results point to a gap in multimodal
embedding models and rerankers for complex retrieval over long-form
video.

\section{Generation Task}
\begin{table*}[ht]
\centering
\small
\resizebox{\textwidth}{!}{%
\begin{tabular}{@{}clccccccc@{}}
\toprule
\textbf{Rank} & \textbf{System}
	& \textbf{Human} & \textbf{Best Votes} & \textbf{Best \%}
	& \textbf{InfoP} & \textbf{InfoR} & \textbf{CiteP} & \textbf{CiteR} \\
\midrule
-- & \textit{CAG (baseline)}
	& \textit{3.088} & \textit{1} & \textit{1.8\%} & \textit{0.764} & \textit{0.410} & \textit{0.617} & \textit{0.228} \\
\midrule
16 & CRAFT-01
	& 2.246 & 0 & 0.0\% & 0.695 & 0.389 & 0.065 & 0.000 \\
15 & CRAFT-02
	& 2.281 & 0 & 0.0\% & 0.627 & 0.304 & 0.530 & 0.198 \\
14 & C2F-RAG-01
	& 2.456 & 6 & 10.5\% & 0.557 & 0.466 & 0.452 & 0.349 \\
13 & C2F-RAG-02
	& 2.526 & 2 & 3.5\% & 0.584 & 0.450 & 0.479 & 0.347 \\
12 & CRAFT-03
	& 2.542 & 0 & 0.0\% & 0.609 & 0.304 & 0.509 & 0.204 \\
11 & TRACE-02
	& 2.579 & 0 & 0.0\% & 0.621 & 0.550 & 0.536 & 0.500 \\
10 & TRACE-01
	& 2.579 & 0 & 0.0\% & 0.599 & 0.564 & 0.515 & 0.498 \\
9 & MARQUIS-Bullet
	& 2.667 & 0 & 0.0\% & 0.711 & 0.394 & 0.604 & 0.237 \\
8 & TRACE-03
	& 2.702 & 2 & 3.5\% & 0.599 & 0.553 & 0.507 & 0.503 \\
7 & MARQUIS-ss-qa-base
	& 3.070 & 6 & 10.5\% & 0.331 & 0.306 & 0.277 & 0.281 \\
6 & MARQUIS-Ginger
	& 3.123 & 6 & 10.5\% & 0.776 & 0.404 & 0.643 & 0.226 \\
5 & MARQUIS-RLM
	& 3.298 & 3 & 5.3\% & 0.708 & 0.385 & 0.592 & 0.272 \\
4 & MARQUIS-ss-qa-ginger
	& 3.421 & 10 & 17.5\% & 0.544 & 0.324 & 0.326 & 0.238 \\
3 & MARQUIS-iter-qa-ginger
	& 3.694 & 5 & 8.8\% & 0.345 & 0.290 & 0.257 & 0.226 \\
2 & TRACE-04
	& 3.825 & 8 & 14.0\% & 0.640 & 0.483 & 0.498 & 0.405 \\
1 & MARQUIS-iter-qa-base
	& 3.833 & 8 & 14.0\% & 0.347 & 0.313 & 0.268 & 0.258 \\
\bottomrule
\end{tabular}%
}
\caption{MAGMaR 2026 --- Oracle generation track leaderboard (sorted by human score). Baseline shown in italics above the rule. 4 teams submitted systems. All teams had at least one system which earned best votes.}
\label{tab:oracle}
\end{table*}

\paragraph{Task.}
The generation task requires systems to produce a cited, persona-consistent article grounded in a set of relevant videos. In the \textit{oracle} setting, systems receive the ground-truth relevant videos for each query, isolating generation quality from upstream retrieval variance. Generated articles are evaluated both by human annotators and by MiRAGE \cite{martin2026seeingmirageevaluatingmultimodal}, an automatic framework that measures factuality, information coverage, groundedness, and citation attribution. Each MiRAGE entailment judgment is produced by CLUE \cite{zhang2026unifiedmultimodaluncertaininference}, a calibrated multimodal uncertain-inference model. Full results are in \autoref{tab:oracle}.
\subsection{Systems}
We provide Collaborative Article Generation (CAG) \cite{martin2025wikivideoarticlegenerationmultiple} as a baseline with Qwen3.5 as the backbone \cite{qwen3.5}. CAG achieves a human score of 3.088 with strong precision on both information (0.76) and citation (0.62) metrics, but low recall.

Four teams submitted 16 predictions to the leaderboard: TRACE \cite{yan2026traceevidencegroundingguidedmultivideo}, CRAFT \cite{bhosale2026craftcriticrefinedadaptivekeyframe}, C2F-RAG \cite{dai26}, and MARQUIS \cite{chakraborty2026marquisthreestagepipelinevideo}. TRACE builds a structured text timeline for each video via OCR and object detection, then uses a text-only LLM to localize relevant moments before passing selected frames to a vision-language model for claim generation. CRAFT generates atomic claims from query-conditioned keyframes and ASR, then iteratively verifies and repairs each claim through a hybrid critic loop before merging citations. C2F-RAG applies its retrieval pipeline's prompt sculpting mechanism to constrain generation to the persona and enforce chunk-level temporal grounding. MARQUIS explores multiple generation strategies over a shared evidence extraction layer, including direct QA, clustering-based summarization, and an RLM controller that iteratively gathers and curates evidence before writing.

\subsection{Results}
The generation results (\autoref{tab:oracle}) show several patterns across the submissions

\paragraph{Human preference vs.\ automatic metrics.} The two top-ranked systems by human score, MARQUIS-iter-qa-base (3.833) and TRACE-04 (3.825), achieve relatively modest MiRAGE precision and recall compared to lower-ranked systems. Conversely, the CAG baseline and MARQUIS-Ginger achieve the highest InfoP (0.764 and 0.776) and CiteP (0.617 and 0.643), yet rank 6th and below by human score. The gap comes mainly from how QA-based systems handle unanswerable sub-questions: when the video evidence is insufficient, these systems refuse to answer rather than hallucinate, and human annotators reward this. MiRAGE, however, scores refusals as missing information and penalizes recall. So the systems human annotators prefer most, precisely because they acknowledge gaps rather than fabricate claims, are the ones automatic evaluation penalizes most.
\paragraph{QA-based generation.} The MARQUIS QA variants dominate the top of the human leaderboard, holding three of the top four positions. These systems decompose the query into sub-questions, answer each from the video evidence, and synthesize the results into an article. Annotators find these outputs more complete and easier to evaluate, likely because the question-answer structure gives the article a natural organization. The gap between iterative and single-shot QA variants (3.833 vs.\ 3.070) suggests that iterating over the evidence, asking follow-up questions and refining answers, produces noticeably better articles.
\paragraph{Evidence verification.} Systems with explicit claim verification show stronger citation precision. CRAFT-03 achieves a CiteP of 0.509 despite ranking 12th overall, and MARQUIS-Ginger reaches 0.643 with its CLUE-based calibration filtering. The TRACE submissions achieve the most balanced precision-recall profiles across both information and citation metrics: TRACE-04 reaches 0.640 InfoP / 0.483 InfoR and 0.498 CiteP / 0.405 CiteR, the strongest combined recall among top-ranked systems.

The shared task results point to several open problems. The clearest is the disconnect between human preference and automatic metrics: systems that abstain when evidence is insufficient are rewarded by annotators but penalized by MiRAGE, which treats all omissions as precision failures. Evaluation frameworks that account for appropriate refusal and persona-relevant (not merely factual) coverage are a clear next step. On the modeling side, no submitted system uses task-specific fine-tuning; the strong performance of QA-based approaches suggests that models trained for grounded multi-video question answering with persona conditioning could improve results. Finally, all current systems convert videos to text proxies before reasoning, discarding visual information that captions cannot capture. Architectures for direct multi-video visual reasoning are needed to close this gap.

\section{Conclusion}
The shared task at the second workshop on Multimodal Augmented Generation via Multimodal Retrieval (MAGMaR 2026) attracted two retrieval teams and four generation teams. Text-based reasoning dominates both tracks: C2F-RAG achieves the highest retrieval scores by indexing summaries and captions rather than operating on video directly, and every generation system converts videos to text before reasoning. Given this shared substrate, inference-time reasoning proves more important than the choice of first-stage model. Reranking transforms OmniEmbed from the weakest retrieval baseline into a competitive first stage, and iterative QA generation produces noticeably better articles than single-shot variants. On the generation side, human preference and automatic metrics also diverge: systems that appropriately abstain when evidence is insufficient are rewarded by annotators but penalized by MiRAGE, which treats all omissions as failures.

These results point to several directions for future work. Evaluation frameworks need to account for appropriate refusal and persona-relevant coverage beyond factual completeness. No submitted system uses task-specific fine-tuning in either track, and the strong performance of QA-based generation suggests that models trained for grounded, persona-conditioned multi-video question answering could improve further. All current systems also discard visual information by converting videos to text before reasoning; architectures for direct multi-video visual reasoning remain unexplored.

\bibliography{custom}

@misc{bhosale2026craftcriticrefinedadaptivekeyframe,
      title={CRAFT: Critic-Refined Adaptive Key-Frame Targeting for Multimodal Video Question Answering}, 
      author={Mahesh Bhosale and Abdul Wasi and Vishvesh Trivedi and Pengyu Yan and Akhil Gorugantu and David Doermann},
      year={2026},
      eprint={2605.19075},
      archivePrefix={arXiv},
      primaryClass={cs.CV},
      url={https://arxiv.org/abs/2605.19075}, 
}

@misc{yan2026traceevidencegroundingguidedmultivideo,
      title={TRACE: Evidence Grounding-Guided Multi-Video Event Understanding and Claim Generation}, 
      author={Pengyu Yan and Akhil Gorugantu and Mahesh Bhosale and Abdul Wasi and Vishvesh Trivedi and David Doermann},
      year={2026},
      eprint={2605.16740},
      archivePrefix={arXiv},
      primaryClass={cs.CV},
      url={https://arxiv.org/abs/2605.16740}, 
}

@misc{chen2025m3embeddingmultilingualitymultifunctionalitymultigranularity,
      title={M3-Embedding: Multi-Linguality, Multi-Functionality, Multi-Granularity Text Embeddings Through Self-Knowledge Distillation}, 
      author={Jianlv Chen and Shitao Xiao and Peitian Zhang and Kun Luo and Defu Lian and Zheng Liu},
      year={2025},
      eprint={2402.03216},
      archivePrefix={arXiv},
      primaryClass={cs.CL},
      url={https://arxiv.org/abs/2402.03216}, 
}

@misc{qwen3.5,
    title  = {{Qwen3.5}: Towards Native Multimodal Agents},
    author = {{Qwen Team}},
    month  = {February},
    year   = {2026},
    url    = {https://qwen.ai/blog?id=qwen3.5}
}

@misc{zhang2026unifiedmultimodaluncertaininference,
      title={Unified Multimodal Uncertain Inference}, 
      author={Dengjia Zhang and Alexander Martin and William Jurayj and Kenton Murray and Benjamin Van Durme and Reno Kriz},
      year={2026},
      eprint={2604.08701},
      archivePrefix={arXiv},
      primaryClass={cs.CV},
      url={https://arxiv.org/abs/2604.08701}, 
}

@misc{martin2026seeingmirageevaluatingmultimodal,
      title={Seeing Through the MiRAGE: Evaluating Multimodal Retrieval Augmented Generation}, 
      author={Alexander Martin and William Walden and Reno Kriz and Dengjia Zhang and Kate Sanders and Eugene Yang and Chihsheng Jin and Benjamin Van Durme},
      year={2026},
      eprint={2510.24870},
      archivePrefix={arXiv},
      primaryClass={cs.CL},
      url={https://arxiv.org/abs/2510.24870}, 
}

@misc{kriz2025multivent20massivemultilingual,
      title={MultiVENT 2.0: A Massive Multilingual Benchmark for Event-Centric Video Retrieval}, 
      author={Reno Kriz and Kate Sanders and David Etter and Kenton Murray and Cameron Carpenter and Kelly Van Ochten and Hannah Recknor and Jimena Guallar-Blasco and Alexander Martin and Ronald Colaianni and Nolan King and Eugene Yang and Benjamin Van Durme},
      year={2025},
      eprint={2410.11619},
      archivePrefix={arXiv},
      primaryClass={cs.CV},
      url={https://arxiv.org/abs/2410.11619}, 
}

@inproceedings{macavaney2022,
  author       = {Sean MacAvaney and
                  Craig Macdonald and
                  Iadh Ounis},
  editor       = {Matthias Hagen and
                  Suzan Verberne and
                  Craig Macdonald and
                  Christin Seifert and
                  Krisztian Balog and
                  Kjetil N{\o}rv{\aa}g and
                  Vinay Setty},
  title        = {Streamlining Evaluation with ir-measures},
  booktitle    = {Advances in Information Retrieval - 44th European Conference on {IR}
                  Research, {ECIR} 2022, Stavanger, Norway, April 10-14, 2022, Proceedings,
                  Part {II}},
  series       = {Lecture Notes in Computer Science},
  volume       = {13186},
  pages        = {305--310},
  publisher    = {Springer},
  year         = {2022},
  url          = {https://doi.org/10.1007/978-3-030-99739-7\_38},
  doi          = {10.1007/978-3-030-99739-7\_38},
  bibsource    = {dblp computer science bibliography, https://dblp.org}
}

@misc{mixedbread2026wholembed,
	title = {Beyond the Limit: Introduce Mixedbread Wholembed v3},
	author = {{Mixedbread Team}},
	year = {2026},
	month = mar,
	howpublished = {Blog post},
	url = {https://mixedbread.com/blog/wholembed-v3}
}

@misc{skow2026rankvideoreasoningrerankingtexttovideo,
      title={RANKVIDEO: Reasoning Reranking for Text-to-Video Retrieval}, 
      author={Tyler Skow and Alexander Martin and Benjamin Van Durme and Rama Chellappa and Reno Kriz},
      year={2026},
      eprint={2602.02444},
      archivePrefix={arXiv},
      primaryClass={cs.IR},
      url={https://arxiv.org/abs/2602.02444}, 
}

@misc{ma2025tevatron20unifieddocument,
      title={Tevatron 2.0: Unified Document Retrieval Toolkit across Scale, Language, and Modality}, 
      author={Xueguang Ma and Luyu Gao and Shengyao Zhuang and Jiaqi Samantha Zhan and Jamie Callan and Jimmy Lin},
      year={2025},
      eprint={2505.02466},
      archivePrefix={arXiv},
      primaryClass={cs.IR},
      url={https://arxiv.org/abs/2505.02466}, 
}

@article{dai26,
    author    = {Jiaxin Dai and Zehang Wei and Jiamin Yan and Xiang Xiang},
    title     = {Decoupling Semantics and Logic: A Training-Free Coarse-to-Fine Pipeline for Video Retrieval-Augmented Generation},
    journal   = {Arxiv preprint},
    volume    = {2606.07924},
    year      = {2026},
    url       = {https://arxiv.org/abs/2606.07924}
}

@misc{lawrie2025overviewtrec2024neuclir,
      title={Overview of the TREC 2024 NeuCLIR Track}, 
      author={Dawn Lawrie and Sean MacAvaney and James Mayfield and Paul McNamee and Douglas W. Oard and Luca Soldaini and Eugene Yang},
      year={2025},
      eprint={2509.14355},
      archivePrefix={arXiv},
      primaryClass={cs.IR},
      url={https://arxiv.org/abs/2509.14355}, 
}

@misc{martin2025wikivideoarticlegenerationmultiple,
      title={WikiVideo: Article Generation from Multiple Videos}, 
      author={Alexander Martin and Reno Kriz and William Gantt Walden and Kate Sanders and Hannah Recknor and Eugene Yang and Francis Ferraro and Benjamin Van Durme},
      year={2025},
      eprint={2504.00939},
      archivePrefix={arXiv},
      primaryClass={cs.CV},
      url={https://arxiv.org/abs/2504.00939}, 
}

@misc{chakraborty2026marquisthreestagepipelinevideo,
      title={MARQUIS: A Three-Stage Pipeline for Video Retrieval-Augmented Generation}, 
      author={Debashish Chakraborty and Dengjia Zhang and Jialiang Jin and Hanting Liu and Katherine Guerrerio and Hanxiang Qin and Tyler Skow and Alexander Martin and Reno Kriz and Benjamin Van Durme},
      year={2026},
      eprint={2605.17640},
      archivePrefix={arXiv},
      primaryClass={cs.IR},
      url={https://arxiv.org/abs/2605.17640}, 
}

\appendix

\section{\textsc{WikiVideo25} Dataset}
\label{append:data}
\textsc{WikiVideo25} builds on \textsc{WikiVideo} \cite{martin2025wikivideoarticlegenerationmultiple}, which contains 57 English topics from 2015--2023, with $\sim$7 videos per topic. It has several limitations: it is English-only, its articles are written top-down from preexisting Wikipedia lead sections, its videos are mostly professional news broadcasts, and its queries are underspecified (e.g., ``tell me about [event name]'') and do not reflect realistic information-seeking behavior \cite{lawrie2025overviewtrec2024neuclir}. Because the topics are well-known events, the relevant information often already exists in pretrained model weights. In some cases, simply quoting the Wikipedia lead section outperforms human-written responses. \textsc{WikiVideo25} targets each of these problems.

\paragraph{Queries and Personas.}
Each topic comes with a persona and a query (two per topic; see \autoref{fig:query-example}). The persona is a detailed background for the query provider. It constrains what counts as relevant evidence, so systems must reason about which parts of the videos matter for a specific professional need, not just surface everything about the event. Each query is a complex information need written in one of two ways: \textit{biased}, where the author had watched the videos and knows what the collection contains, and \textit{unbiased}, where the author had only seen video titles and writes a more natural request without that knowledge. Unbiased queries may ask for information absent from the collection, so they also test whether systems can acknowledge gaps in the evidence instead of hallucinating an answer.

\paragraph{Video-Centric Reports.}
\textsc{WikiVideo}'s articles were written top-down: annotators began with the Wikipedia lead section for each event and removed claims not supported by the videos. The ground-truth articles were therefore anchored to an external text source, not to the video content itself, and in some cases quoting the Wikipedia lead outperformed human-written responses. \textsc{WikiVideo25} reverses this. Annotators first write atomic claims grounded in each video's audiovisual content, then combine them into higher-level inferences that form the final article. This bottom-up annotation produces reports more faithful to the video content and reduces the advantage of parametric knowledge, since the ground truth is no longer derivable from a well-known Wikipedia article.

\paragraph{Knowledge cutoff.} 
All events in \textsc{WikiVideo25} take place in 2025, after the training cutoffs of the models used in submitted systems. Because none of the event-level facts exist in these models' parametric knowledge, systems cannot shortcut the task by recalling memorized information and must ground their outputs in the retrieved video evidence. This is resolves the quoting issue noted above from \textsc{WikiVideo}, whose topics were well-known events with extensive coverage in pretraining data. Using events beyond any model's knowledge cutoff removes this issue. It also means the MultiVENT 2.0 distractor set, which covers events only through 2023, contains no true positives for any \textsc{WikiVideo25} query. The result is a cleaner evaluation of both retrieval and generation: retrieval systems must identify relevant videos rather than match familiar event descriptions, and generation systems must synthesize articles from audiovisual content rather than parametric knowledge.

\paragraph{Raw Videos.}
The original dataset is mostly professional news broadcasts, where a third-party narrator describes the event to the viewer. For these, an audio transcript alone is often enough to produce a reasonable article. \textsc{WikiVideo25} shifts toward raw and ``diet raw'' footage: videos captured by participants or bystanders without professional narration or post-production editing. These videos are more ambiguous and force models to infer what is happening from low-level audiovisual cues. There is no narrator explaining the scene.

\paragraph{Multilingual Videos.}
\textsc{WikiVideo25} adds two topics with Mandarin-language videos and one mixed Nepali--English topic. Systems must now process non-English audio and OCR at both retrieval and generation. In the mixed-language case, they must also reason across videos in different languages about the same event.

\end{document}